\documentclass[10pt,twocolumn,letterpaper]{article}

\usepackage[pagenumbers]{iccv}

\definecolor{iccvblue}{rgb}{0.21,0.49,0.74}
\usepackage[pagebackref,breaklinks,colorlinks,allcolors=iccvblue]{hyperref}
\usepackage{amssymb}
\usepackage{makecell}

\title{RouteExtract: A Modular Pipeline for Extracting Routes from Paper Maps}

\author{Bjoern Kremser\\
Technical University of Munich\\
Boltzmannstr. 3, 85748 Garching, Germany\\
The University of Tokyo\\
7-3-1 Hongo, Bunkyo-ku, Tokyo, Japan\\
{\tt\small bjoern.kremser@tum.de}
\and
Yusuke Matsui\\
The University of Tokyo\\
7-3-1 Hongo, Bunkyo-ku, Tokyo, Japan\\
{\tt\small matsui@hal.t.u-tokyo.ac.jp}
}

\begin{document}
\maketitle

\begin{abstract}
Paper maps remain widely used for hiking and sightseeing because they contain curated trails and locally relevant annotations that are often missing from digital navigation applications such as Google Maps. We propose a pipeline to extract navigable trails from scanned maps, enabling their use in GPS-based navigation. Our method combines georeferencing, U-Net-based binary segmentation, graph construction, and an iterative refinement procedure using a routing engine. We evaluate the full end-to-end pipeline as well as individual components, showing that the approach can robustly recover trail networks from diverse map styles and generate GPS routes suitable for practical use.
\end{abstract}

\section{Introduction}

Although digital navigation tools are the primary resource for route planning in everyday life, paper maps remain widely used in domains such as hiking and sightseeing. These maps often feature curated, hand-designed trails, such as multiple ascent routes of varying difficulty or sightseeing loops connecting key landmarks, that reflect local knowledge and design intent. Algorithmic routing from start to goal lacks this context and may overlook meaningful or recommended paths.

\begin{figure}[t]
    \centering
    \begin{subfigure}[t]{\linewidth}
        \centering
        \includegraphics[width=\linewidth]{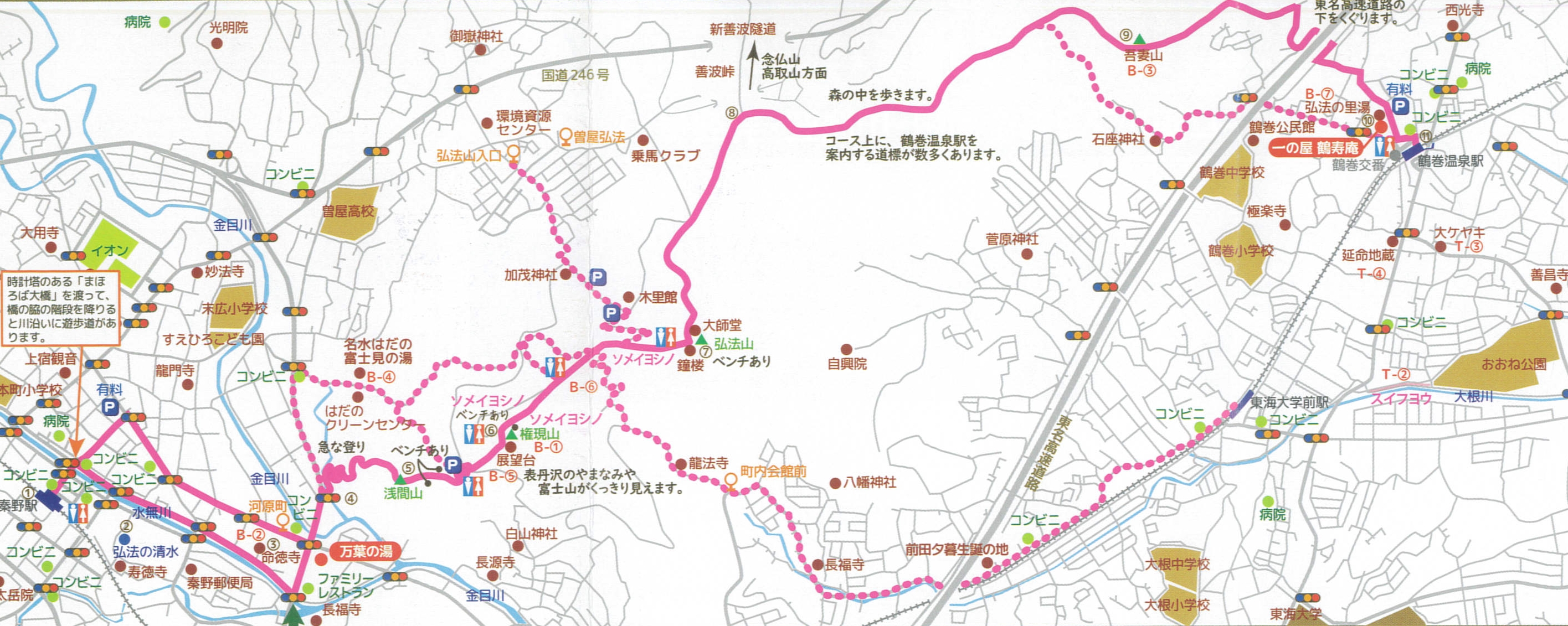}
        \caption{Input scanned map}
    \end{subfigure}
    \hfill
    \begin{subfigure}[t]{\linewidth}
        \centering
        \includegraphics[width=\linewidth]{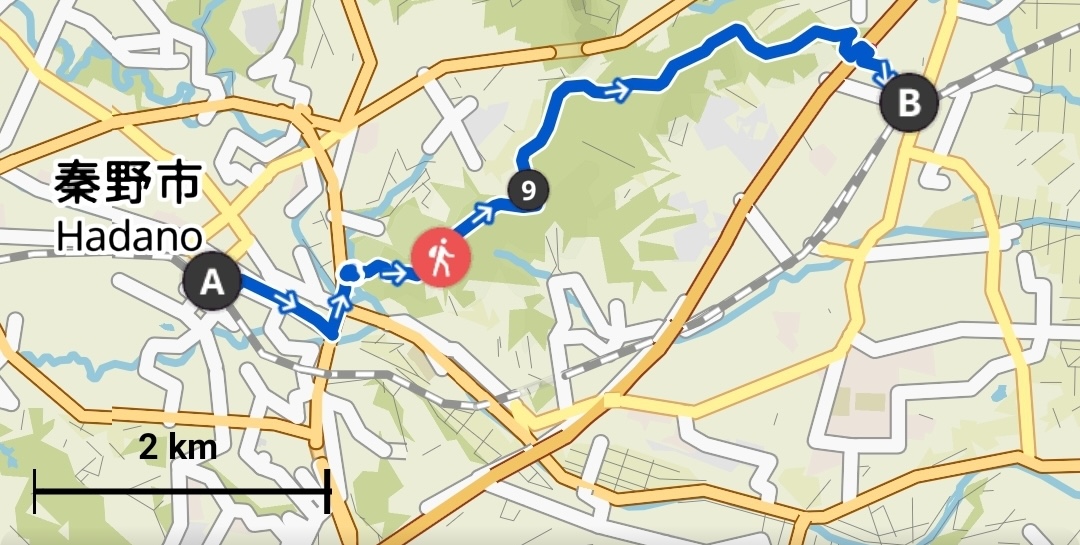}
        \caption{Generated GPX route opened in smartphone navigation app Komoot}
    \end{subfigure}
    \caption{Overview of the pipeline input and output. A scanned hiking map (a) is processed to extract a routable GPX track (b).}
    \label{fig:overview_preview}
\end{figure}

We propose an automated pipeline to extract such curated trails from scanned maps, combining their local knowledge with the convenience and precision of GPS-based navigation. Our system segments scanned map images, constructs a graph representation of the trail network, and generates navigable routes using an iterative refinement procedure. The output is encoded in the widely adopted GPX (GPS Exchange) format, enabling direct integration with modern GPS devices and apps such as Google Maps. \autoref{fig:overview_preview} shows an example of a scanned map input and the corresponding generated GPX route.

The main contributions of this work are as follows:
\begin{itemize}
    \item A modular pipeline for extracting curated GPX routes from scanned hiking and sightseeing maps.
    \item A graph-based representation of trail structure supporting robust route generation with complex trail layouts.
    \item A quantitative evaluation demonstrating the effectiveness of the approach across diverse map styles.
\end{itemize}

\begin{figure*}[t]
    \centering
    \begin{subfigure}[t]{0.23\textwidth}
        \centering
        \includegraphics[width=\linewidth]{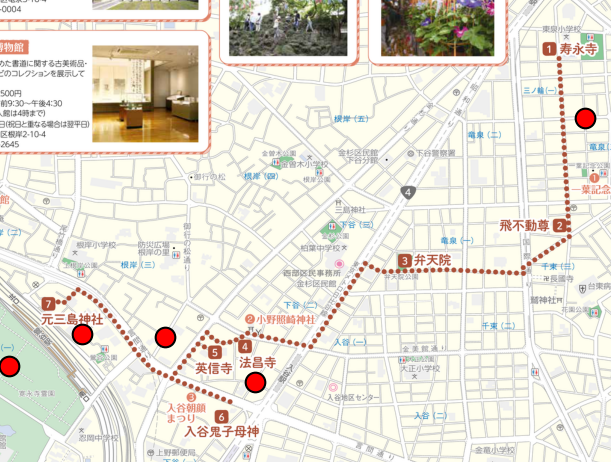}
        \caption{Ground control points (GCPs) marked on input map}
    \end{subfigure}
    \hfill
    \begin{subfigure}[t]{0.23\textwidth}
        \centering
        \includegraphics[width=\linewidth]{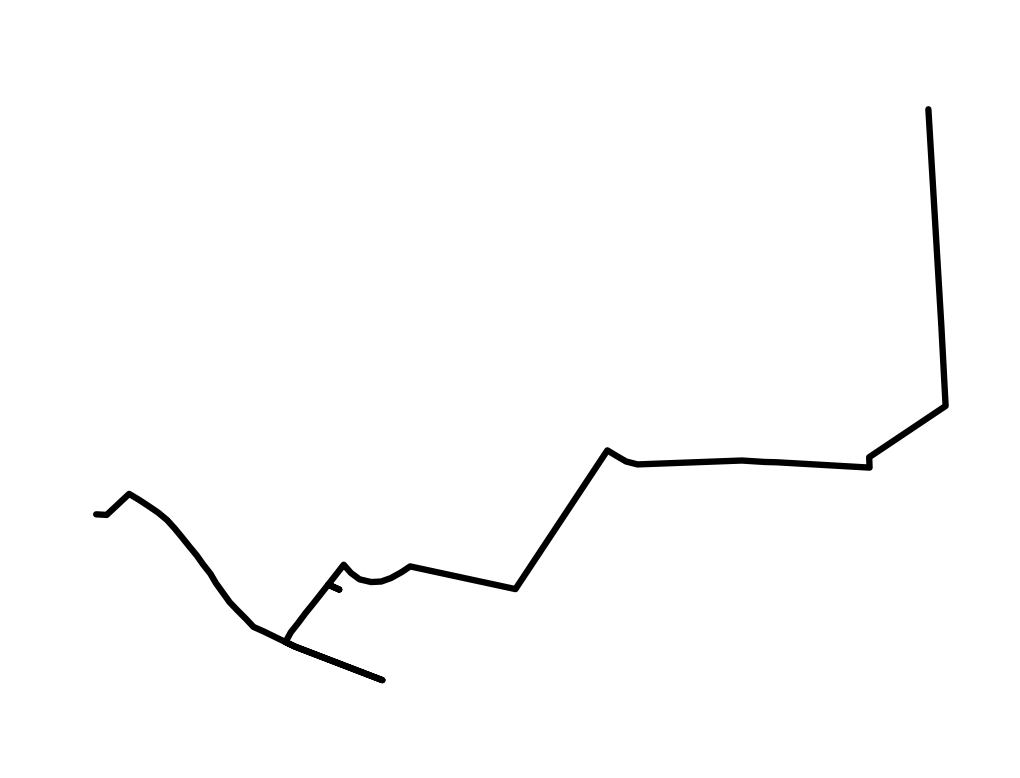}
        \caption{Extracted trail mask}
    \end{subfigure}
    \hfill
    \begin{subfigure}[t]{0.23\textwidth}
        \centering
        \includegraphics[width=\linewidth]{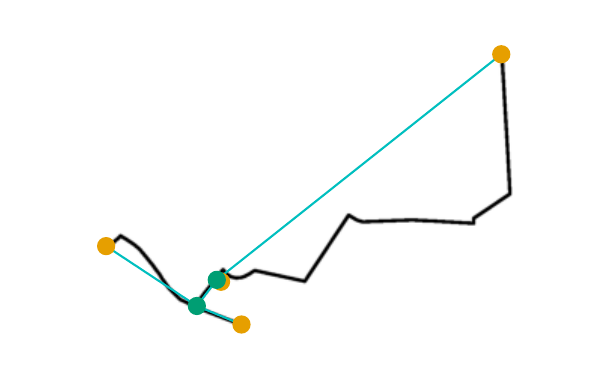}
        \caption{Graph representation of extracted trail}
    \end{subfigure}
    \hfill
    \begin{subfigure}[t]{0.23\textwidth}
        \centering
        \includegraphics[width=\linewidth]{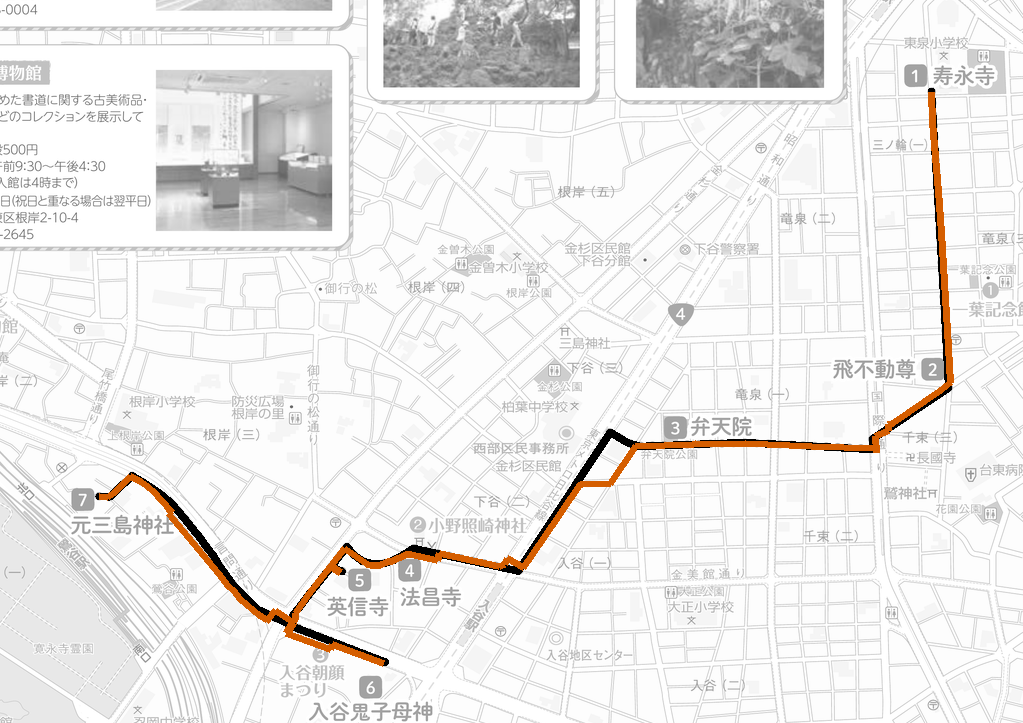}
        \caption{Generated route (orange) overlaid on the input map, with the ground truth trail shown in black}
    \end{subfigure}
    \caption{Overview of the pipeline: (a) input map with manually specified GCPs, (b) image segmentation to extract a trail, (c) construction of the graph representation, and (d) final routable GPX track.}
    \label{fig:pipeline-overview}
\end{figure*}

To the best of our knowledge, no prior work has directly addressed the problem of extracting curated trails from scanned hiking or sightseeing maps. Related research has primarily focused on (i) digitizing historical maps \cite{automatic-georeferencing-survey, Affek2013375} and extracting specific features from them \cite{Uhl20206978}, (ii) extracting road networks from aerial or satellite imagery \cite{Bastani20184720, Zhu2021353}, and (iii) using pixel adjacency graphs for image processing \cite{malmberg-graph-image-segmentation}. In contrast to prior research, we focus on curated trails overlaid on maps, which indicate recommended routes rather than strictly geographic information. These trails are typically drawn in a distinct style on top of existing map features to be easily recognized by the reader.

\section{Method}
We propose a pipeline consisting of four main steps: georeferencing the input map, segmenting the map to extract trails, building a graph representation of the trails, and finally generating a routable GPX track (see \autoref{fig:pipeline-overview}).

Before describing each component, we first clarify our definition of a \textit{trail} in this work. A trail is a path drawn on a map, typically with a distinct color or stroke style, indicating a route between two or more points of interest with a clear start and end position. On the rasterized map image, trails occupy a region of nonzero width, even though they conceptually represent a connected sequence of points. Trails can split or rejoin, and individual sections may be traversed in multiple directions. In later steps, we represent trails as a graph: junctions and dead ends correspond to nodes, and the portions of trail between two such nodes are referred to as \textit{segments}.

\subsection{Georeferencing}\label{subsec:georeferencing}

Our system involves two types of spatial data: \textit{raster data}, such as scanned maps and trail masks, defined in pixel-based image coordinates, and \textit{vector data}, such as OpenStreetMap (OSM) data or GPX tracks, defined in geographic coordinates (latitude and longitude). These coordinate systems are inherently incompatible, necessitating a transformation between them. This transformation is established through a process known as \textit{georeferencing}~\cite{georef}.

Let the input image be denoted by \( \mathbf{I} \in \mathbb{R}^{H \times W \times 3} \), where \( H \) and \( W \) are the image height and width, respectively. We identify a set of \( N \) corresponding ground control points (GCPs) between the image and the geographic domain:

\begin{equation}
\small
\{\mathbf{x}_i\}_{i=1}^N, \quad \mathbf{x}_i \in [1, H] \times [1, W] \quad \text{(image coordinates)}
\end{equation}
\begin{equation}
\{\mathbf{y}_i\}_{i=1}^N, \quad \mathbf{y}_i \in \mathbb{R}^2 \quad \text{(latitude, longitude)}
\end{equation}

Using these point correspondences, we estimate an affine transformation \( T: \mathbb{R}^2 \rightarrow \mathbb{R}^2 \), parameterized as
\begin{equation}
T(\mathbf{x}) = A \mathbf{x} + \mathbf{t}
\end{equation}
where \( A \in \mathbb{R}^{2 \times 2} \) is a linear transformation matrix and \( \mathbf{t} \in \mathbb{R}^2 \) is a translation vector. Since more than the minimum three GCPs are typically provided, \( T \) is computed via least squares minimization to approximately satisfy \( T(\mathbf{x}_i) \approx \mathbf{y}_i \) for all \( i = 1, \dots, N \). In practice, we observed that providing around 5 to 10 GCPs were sufficient to achieve accurate alignment. The example in \autoref{fig:pipeline-overview} uses 5 GCPs for the computation of \( T \).

The transformation \( T \) enables projecting any image coordinate \( \mathbf{x} \in \mathbb{R}^2 \) into geographic space. Its inverse \( T^{-1} \), which is well-defined for non-degenerate affine transforms, allows projecting geographic coordinates back into image space. This bidirectional mapping is essential for integrating outputs from the raster and vector processing stages of the pipeline.

Identifying GCPs remains a largely manual process, although research has explored automation strategies \cite{automatic-georeferencing-historical, automatic-georeferencing-survey, automatic-georeferencing-topological}. In this work, we assume that the GCPs are manually specified during dataset preparation and not yet generated automatically. They are provided during the route generation step to compute the transformation \( T \) for each map.

\begin{figure*}[t]
    \centering
    \begin{subfigure}[t]{0.3\textwidth}
        \centering
        \includegraphics[width=\linewidth]{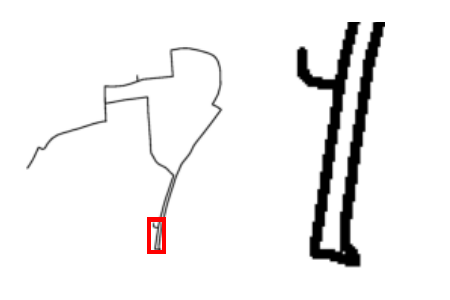}
        \caption{Trail Mask $\mathbf{B}$}
    \end{subfigure}
    \hfill
    \begin{subfigure}[t]{0.3\textwidth}
        \centering
        \includegraphics[width=\linewidth]{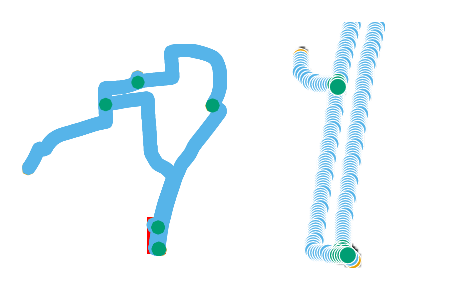}
        \caption{Initial dense graph \( G \)}
    \end{subfigure}
    \hfill
    \begin{subfigure}[t]{0.3\textwidth}
        \centering
        \includegraphics[width=\linewidth]{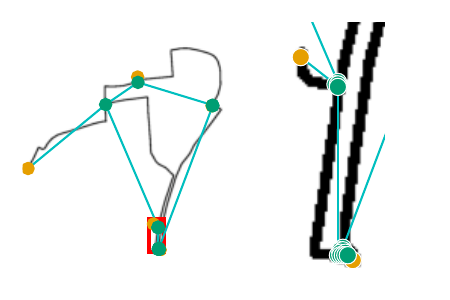}
        \caption{Simplified graph \( G' \) after first linear path contraction}
    \end{subfigure}
    \vspace{0.5em}
    \makebox[0.65\textwidth][c]{
    	\begin{subfigure}[t]{0.3\textwidth}
        	\centering
        	\includegraphics[width=\linewidth]{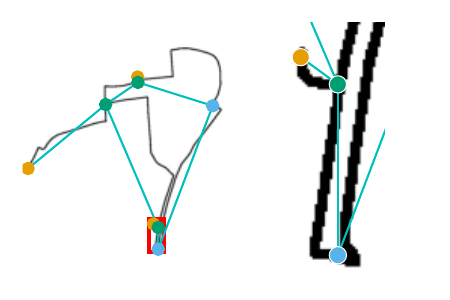}
        	\caption{\( G' \) after collapsing nearby nodes}
    	\end{subfigure}
    	\hfill
    	\begin{subfigure}[t]{0.3\textwidth}
        	\centering
        	\includegraphics[width=\linewidth]{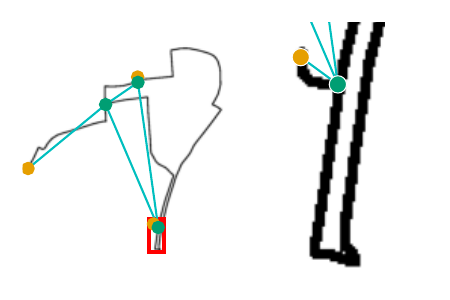}
        	\caption{\( G' \) after second linear path contraction}
    	\end{subfigure}
    }
    \caption{Graph simplification on a typical sightseeing trail map, showing the full graph alongside a zoomed-in portion.
The original trail mask \( \mathbf{B} \) (a) contains a main loop with detours representing points of interest. 
From the skeletonized trail mask, we build an initial dense graph (b), which is simplified in three steps (c-e). If the graph was disconnected, we would connect the components in a fourth step.
The final graph retains only semantically relevant nodes: \textbf{leaf nodes} (yellow) and \textbf{junctions} (green). Nodes with degree 2 are drawn in blue.}
    \label{fig:graph_processing}
\end{figure*}

\subsection{Image Segmentation}\label{subsec:image_segmentation}

A key step in our pipeline is the identification of trail regions in the input map image. We formulate trail detection as a binary segmentation problem, where each pixel is classified as either \textit{trail} or \textit{background}. The model produces a binary mask $\mathbf{B} \in \{0, 1\}^{H \times W}$, where $B_{i,j} = 1$ denotes trail presence at pixel location $(i, j)$, and $B_{i,j} = 0$ otherwise.

A challenge in this task is that maps often contain multiple trails, and the number and appearance of these trails vary across samples. Assigning a unique class label to each trail is impractical, as the class identities are not consistent across maps. However, most maps distinguish trails by color, which we leverage to guide the segmentation process.

To condition the model on the visual characteristics of a specific trail, we use its color as an additional input. The input to the segmentation model consists of the RGB map image \( \mathbf{I} \) stacked with the trail color \( \mathbf{c} \in \mathbb{R}^3 \), broadcast across the spatial dimensions. This results in a 6-channel input tensor \( \mathbf{X} \in \mathbb{R}^{ H \times W \times 6} \), where the additional three channels represent the trail color repeated at every pixel location.
The color vector \( \mathbf{c} \) is obtained manually by sampling a representative RGB value of the trail from the map. For example, in \autoref{fig:pipeline-overview}, the brown trail is associated with the sampled color \( \mathbf{c} = (170, 79, 55) \).

We use a U-Net architecture~\cite{ronneberger2015unet} to perform the segmentation. For each trail in a map, the model receives the corresponding 6-channel input \( \mathbf{X} \) and outputs the binary mask \( \mathbf{B} \). As a result, we can obtain separate binary masks for each trail in a map.

\subsection{Building a Graph Representation of Trails}
\label{subsec:graph}

For simple trails that consist of a single path from start to finish, we can often proceed directly to route generation (see \autoref{subsec:route_generation}). However, more complex trails that include loops, revisited sections, or branching paths require additional preprocessing to ensure the reconstructed route accurately reflects the intended path. To this end, we build a graph representation that captures a semantic structure of key points, such as junctions and dead ends (see \autoref{fig:pipeline-overview}), which provides the foundation for planning a route that covers all trail segments. This comprehensive coverage is particularly important for sightseeing maps, which often start and end at the same location while visiting multiple landmarks along a set route.

We construct this representation by converting the binary trail mask \( \mathbf{B} \in \{0, 1\}^{H \times W} \) into a graph structure that captures the layout of the trail network. As a first step, we skeletonize the binary trail mask:
\begin{equation}
\mathbf{S} = \mathrm{skeletonize}(\mathbf{B}), \quad \mathbf{S} \in \{0, 1\}^{H \times W}.
\end{equation}
Each foreground pixel \((i, j)\) in the skeleton is treated as a graph node. We define the vertex set:
\begin{equation}
V = \left\{ (i, j) \in \{1, \dots, H\} \times \{1, \dots, W\} \mid S_{i, j} = 1 \right\}.
\end{equation}
Edges are added between all 8-connected foreground pixels. That is, for each \( \mathbf{v} = (i, j) \in V \), and each neighboring pixel \( \mathbf{v}' = (i', j') \in V \) such that \( \| \mathbf{v} - \mathbf{v}' \|_\infty = 1 \), we include an undirected edge:
\begin{equation}
E = \left\{ \{ \mathbf{v}, \mathbf{v}' \} \mid \mathbf{v}, \mathbf{v}' \in V,\ \| \mathbf{v} - \mathbf{v}' \|_\infty = 1 \right\}.
\end{equation}

This yields an initial undirected graph \( G = (V, E) \) that densely represents the skeletonized trail structure. Due to its pixel-level resolution, this graph contains many redundant nodes that are not semantically meaningful. To make the graph semantically meaningful and suitable for route planning, we simplify it through a four-stage process (see \autoref{fig:graph_processing}). The first three stages reduce noise and redundancy, while the final stage ensures the graph is a single connected structure, even in the presence of prediction artifacts or map design discontinuities.

\begin{enumerate}
    \item \textbf{Linear path contraction (first pass)}: 
    We simplify linear paths in the graph by removing intermediate nodes with degree 2, i.e. nodes connected to exactly two other nodes. 
    For each such path \( ( \mathbf{v}_1, \dots, \mathbf{v}_k ) \subset V \), we require that the endpoints \( \mathbf{v}_1 \) and \( \mathbf{v}_k \) have degree not equal to 2, while all intermediate nodes satisfy \( \deg(\mathbf{v}_i) = 2 \) for \( 2 \leq i \leq k-1 \). 
    We replace this chain with a single edge between \( \mathbf{v}_1 \) and \( \mathbf{v}_k \).     
    The simplified graph \( G' = (V', E') \) is constructed incrementally: we initialize \( V' = \emptyset \), \( E' = \emptyset \), and for each contracted path, add endpoints \( \mathbf{v}_1 \), \( \mathbf{v}_k \) to \( V' \) and the new edge \( \{ \mathbf{v}_1, \mathbf{v}_k \} \) to \( E' \).

    \item \textbf{Node collapsing}: 
    Minor visual noise in the skeletonization can result in multiple nearby nodes that semantically represent the same junction or endpoint. 
    To remove this redundancy, we cluster nodes in \( V' \) that lie within a small Euclidean distance threshold \( \tau \).

    \item \textbf{Linear path contraction (second pass)}: 
    After collapsing nodes, new linear segments may form. We therefore repeat the first step on the updated graph to eliminate any newly introduced linear paths.
    \item \textbf{Connecting graph components}: There are cases where the graph contains multiple components at this point. This can result from map design elements like overlaid labels or from prediction artifacts that introduce breaks or noise into the trail. 
    To ensure global connectivity, we iteratively connect the closest pair of components by adding an edge between the pair of nodes with minimal Euclidean distance. This process is repeated until a single connected graph remains.  
\end{enumerate}

After these simplification steps, the resulting simplified graph \( G' \) is more compact and semantically structured, containing only the key nodes necessary for reasoning about trail connectivity. In particular, we distinguish:
\begin{itemize}
  \item \textit{Leaf nodes} (degree = 1), which represent trail termini or endpoints of linear segments\footnote{Also referred to as “end nodes” in some contexts. We use “leaf node” to avoid confusion with the route's destination.}.
  \item \textit{Junctions} (degree $\geq$ 3), which indicate branching points or intersections.
\end{itemize}

While the graph primarily consists of these two types, an exception can arise after connecting disconnected components: nodes that were originally leaf nodes (degree = 1) may become degree-2 nodes after gaining a new edge. These are retained despite not strictly fitting into the above categories, as they play an essential role in ensuring the overall connectivity of the graph. We avoid reapplying the simplification step after this stage to preserve such structurally important nodes.

Despite such exceptions, the simplified graph retains a clear semantic structure composed of leaf nodes and junctions. These serve as the key elements for route planning. For instance, leaf nodes (yellow in \autoref{fig:pipeline-overview} and \autoref{fig:graph_processing}) typically represent dead-ends that must be visited and backtracked, while junctions (green) correspond to branching points where the route must split and rejoin.

We perform a breadth-first traversal from the start to the end node that visits all key nodes in \( G' \). This problem is related to the \textit{Traveling Salesman Problem} in that we seek to cover a set of nodes with minimal redundancy \cite{lawler1985tsp}.

After the order of key node visits is determined, we recover the detailed route in the full graph \(G\). For each consecutive pair of key nodes along the simplified path, we identify corresponding nodes in \(G\) and perform a local search to find a connecting path. Intermediate points along these paths are then sampled to construct a dense waypoint sequence, which we use for route generation in \cref{subsec:route_generation}.

\subsection{Iterative Route Generation and Evaluation}
\label{subsec:route_generation}

To generate a routable GPX track, we use an external routing engine, like the open-source GraphHopper, to compute a path that aligns with the trail mask obtained in \cref{subsec:image_segmentation}. We iteratively refine this route by identifying high-error regions and expanding the list of input waypoints for future routing requests.

We begin with an initial GPX route, generated either from a simple start-to-end query or from a list of waypoints obtained from the trail graph (\cref{subsec:graph}). The GPX output consists of points in geographic coordinates, which are projected into image space using the georeferencing transformation \( T^{-1} \) defined in \cref{subsec:georeferencing}.

Let \( \mathcal{R} = \{ \mathbf{r}_1, \dots, \mathbf{r}_n \} \subset \mathbb{R}^2 \) denote the current GPX route points projected into image space. These points define a polyline \( \mathcal{P} \), formed by connecting consecutive points \( \mathbf{r}_i \) and \( \mathbf{r}_{i+1} \). Similarly, let \( \mathcal{T} = \{ \mathbf{t}_1, \dots, \mathbf{t}_m \} \subset \mathbb{R}^2 \) denote the set of foreground trail pixels extracted from the binary mask \( \mathbf{B} \) (see the final route overlay in \autoref{fig:pipeline-overview} for a visual illustration of the projected polyline).

To assess where the generated route diverges from the trail mask, we compare \( \mathcal{T} \) with the polyline \( \mathcal{P} \). We consider two types of high-error regions:
\begin{itemize}
    \item Route points \( \mathbf{r}_i \in \mathcal{R} \) that are far from any trail pixel in \( \mathcal{T} \),
    \item Trail pixels \( \mathbf{t}_j \in \mathcal{T} \) that are far from the route polyline \( \mathcal{P} \). Here, the distance from \( \mathbf{t}_j \) to \( \mathcal{P} \) is defined as the minimum Euclidean distance from \( \mathbf{t}_j \) to any line segment \( [\mathbf{r}_i, \mathbf{r}_{i+1}] \) in the polyline.
\end{itemize}

In the first case, we determine the closest trail pixel to the high-error route point \( \mathbf{r}_i \), denoted \( \mathrm{NN}_{\mathcal{T}}(\mathbf{r}_i) \), and convert it to geographic coordinates using \( T \). This new point \( T(\mathrm{NN}_{\mathcal{T}}(\mathbf{r}_i)) \) is then added as an additional waypoint. In the second case, we directly use the geographic projection \( T(\mathbf{t}_j) \) of the high-error trail pixel \( \mathbf{t}_j \) as a waypoint.

We then perform two separate routing requests: one using the additional waypoint from the route-error case, and one from the trail-error case. We select the route with the lower Chamfer distance to use in the next iteration. Formally, the Chamfer distance between route points \( \mathcal{R} \) and a trail mask \( \mathcal{T} \) is defined as:

{\scriptsize
\begin{equation}
	d_\text{Chamfer}(\mathcal{R}, \mathcal{T}) = 
	\frac{1}{2|\mathcal{R}|} \sum_{\mathbf{r} \in \mathcal{R}} \min_{\mathbf{t} \in \mathcal{T}} \| \mathbf{r} - \mathbf{t} \|_2 +
	\frac{1}{2|\mathcal{T}|} \sum_{\mathbf{t} \in \mathcal{T}} \min_{\mathbf{r} \in \mathcal{R}} \| \mathbf{t} - \mathbf{r} \|_2
\end{equation}
}

This refinement process is repeated until either a maximum number of iterations is reached or no further improvement in Chamfer distance is observed.

% Experiments
\section{Experimental Setup}

We evaluate our system at three levels: two core components in isolation, the complete pipeline end-to-end, and an ablation study measuring the contribution of different parts of our route generation algorithm to final performance.

First, we assess the segmentation quality of predicted trail masks and the performance of route generation when provided with ground truth trail masks.  
Second, we evaluate the full end-to-end pipeline, where predicted masks are used as input to the route generation module.
Finally, our ablation study is designed to answer two key questions:
\begin{itemize}
	\item \textbf{What is the impact of the initial query strategy?}  
	We compare three variants:
		(i) an end-to-end query using the trail’s start and goal positions,  
		(ii) a graph-based query using waypoints extracted as described in \cref{subsec:graph}, and  
		(iii) a hybrid approach that evaluates both and chooses the better-performing option as input to the iterative refinement stage.
	\item \textbf{Does iterative refinement improve results?}  
	We compare the results of using the best-performing start query with and without iterative refinement.
\end{itemize}

\subsection{Dataset and Configuration}

We manually curated a dataset of 50 hiking and sightseeing maps from Japan, comprising a total of 85 trails. Each trail has an associated ground truth trail mask, created in Affinity Photo. Most maps also include a set of GCPs for georeferencing. Scripts for generating the dataset locally will be provided in the \texttt{RouteExtract} GitHub repository.

For model training, we split the dataset into 35 training maps (63 trails), 5 validation maps (9 trails), and 10 test maps (13 trails). All input images and trail color tensors are resized to \(1024 \times 1024\) resolution and normalized. We apply standard data augmentations such as random horizontal/vertical flips and rotations to improve generalization. The U-Net model is trained using binary cross-entropy loss, with early stopping applied if validation loss does not improve for 10 consecutive epochs. The best-performing checkpoint (lowest validation loss) is used for evaluation.

Route generation is performed using a locally hosted GraphHopper instance, configured with the pedestrian profile and based on an OpenStreetMap snapshot of Japan dated 2025-04-21. All experiments were run on a PC with an Intel i7-10870H CPU, 64GB RAM, and an NVIDIA RTX 3080 Laptop GPU with 16GB of VRAM.

\begin{figure}
    \centering
    \begin{subfigure}[t]{0.95\linewidth}
        \includegraphics[width=\textwidth]{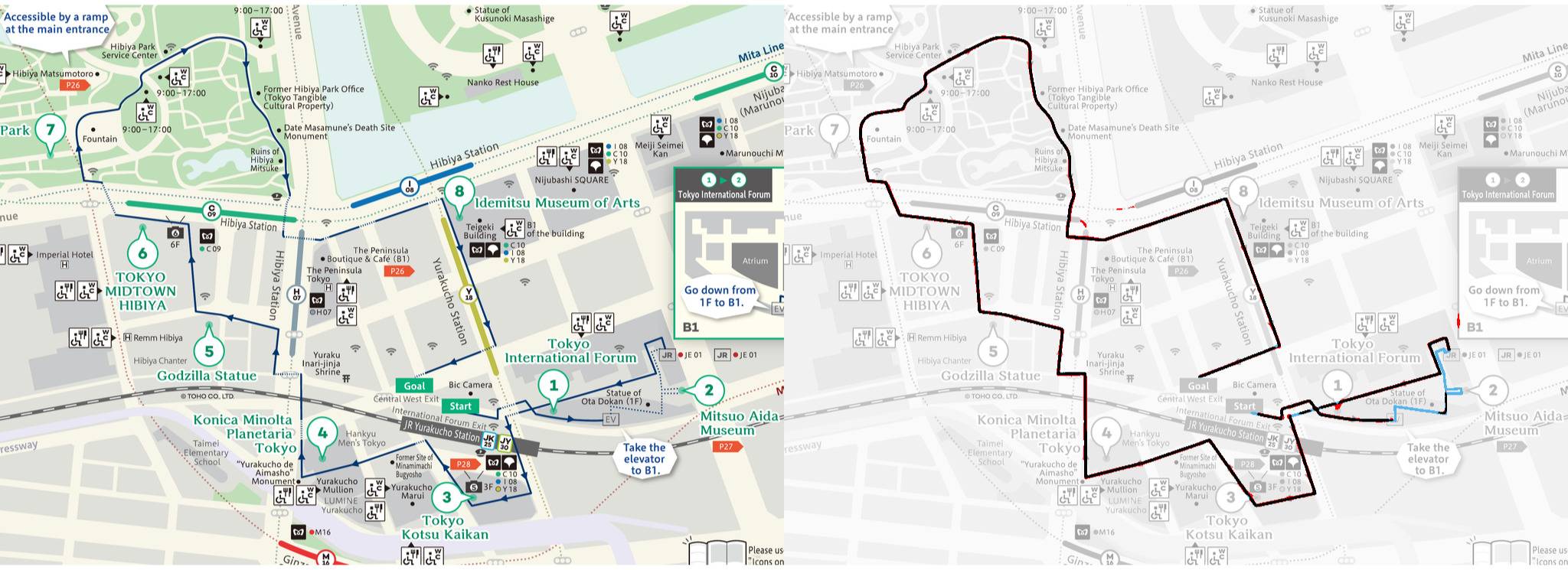}
        \caption{Example with median IoU of 0.76. Correct predictions are black, false positives are \textcolor[HTML]{FF0000}{red}, and missed trail pixels are \textcolor[HTML]{56B4E9}{blue}. Apart from a short, differently styled section, the trail is mostly identified correctly, with only minor noise picked up.}
    \end{subfigure}
    \vspace{0.5em}
    \begin{subfigure}[t]{0.95\linewidth}
        \includegraphics[width=\textwidth]{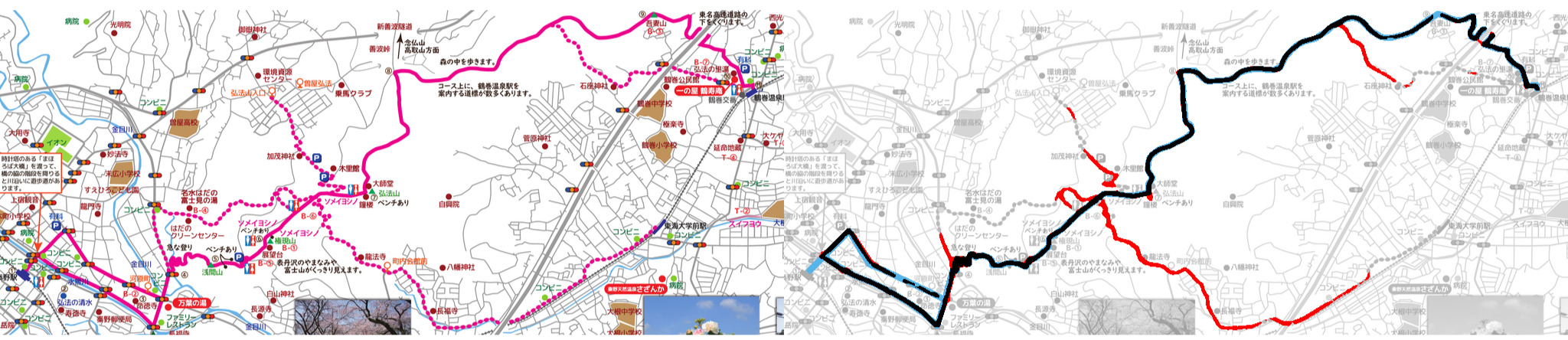}
        \caption{Example with low IoU of 0.55. Dotted lines are incorrectly inferred as part of the trail instead of as alternative paths.}
    \end{subfigure}
    \caption{
Segmentation results on two examples. Left: input map; Right: prediction overlay with color-coded pixel-wise comparison. These cases illustrate how style decisions can affect segmentation quality.
    }
    \label{fig:segmentation_results}
\end{figure}

\subsection{Metrics}

To evaluate segmentation quality, we use the standard Intersection-over-Union (IoU) metric~\cite{iou}. For assessing the accuracy of generated GPX routes, we use the Chamfer distance~\cite{barrow77chamfer}, which is robust to minor spatial misalignments between predicted routes and trail masks. Additionally, we report the distances from GPX points to the trail mask and vice versa, providing a more detailed picture of local alignment errors.

For all distance-based metrics, both the GPX routes and trail masks are projected into the EPSG:32654 coordinate reference system (UTM Zone 54N), enabling distance computations in meters within the spatial bounds of the dataset.
% Results
\section{Results}

\subsection{Segmentation Quality}

The segmentation model achieved a median IoU of 0.763, with an interquartile range of 0.130. Notably, no outliers were detected, suggesting consistent performance across all trail samples. Qualitatively, most predicted masks closely resemble their ground truth counterparts.

The lowest-performing trails tend to appear on maps where visual ambiguity hinders segmentation. For example, the worst-case trail (IoU = 0.548) uses stroke style rather than color to distinguish alternate paths which our model is not designed to correctly interpret, as seen in \cref{fig:segmentation_results}. Other low-IoU cases also include design elements that are similarly colored and styled to the trails. For example, one map has a red trail while an unrelated train line is also drawn in a similar red color.

Overall, segmentation performance is robust, but susceptible to design-induced ambiguity in the source maps.

\subsection{Route Generation in Isolation}

To isolate the performance of the route generation algorithm, we use clean ground truth trail masks as input, thereby removing the effect of segmentation noise.

The median Chamfer distance between the generated GPX routes and the ground truth trail masks is 13.04 meters, with a standard deviation of 11.40 meters. The highest observed Chamfer distance in the test set was 39.42 meters. These results indicate that the generated routes generally follow the intended trails closely, with only minor deviations in most cases.

\begin{figure}
    \centering
    \includegraphics[width=0.95\linewidth]{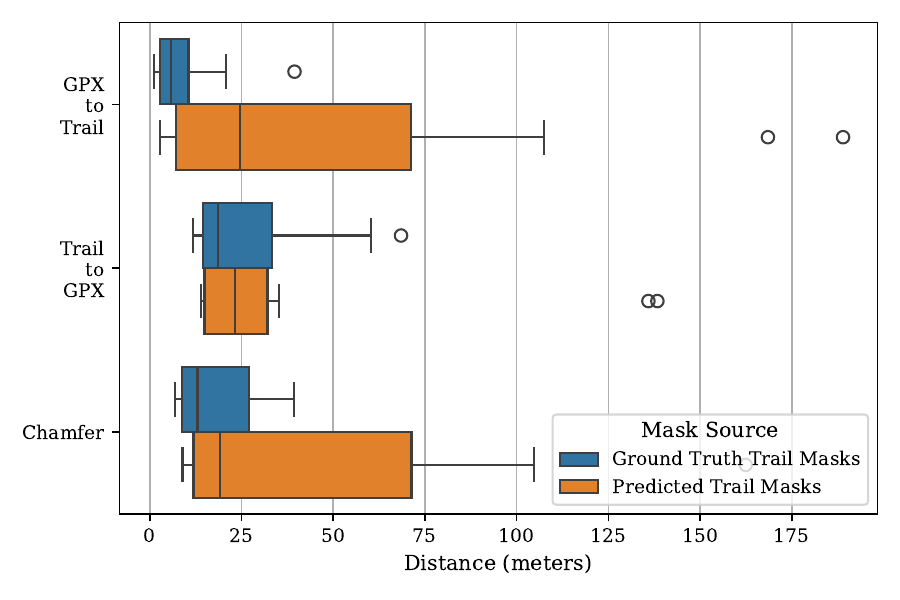}
    \caption{
    Distance-based accuracy of extracted GPX routes using ground truth (blue) and predicted (orange) trail masks as inputs to the route generation algorithm. 
    Performance degrades when using predicted masks, with significantly higher GPX-to-trail deviations. 
    }
    \label{fig:distances}
\end{figure}

\begin{figure*}[t]
    \centering
    \begin{subfigure}[t]{0.23\textwidth}
        \includegraphics[width=\linewidth]{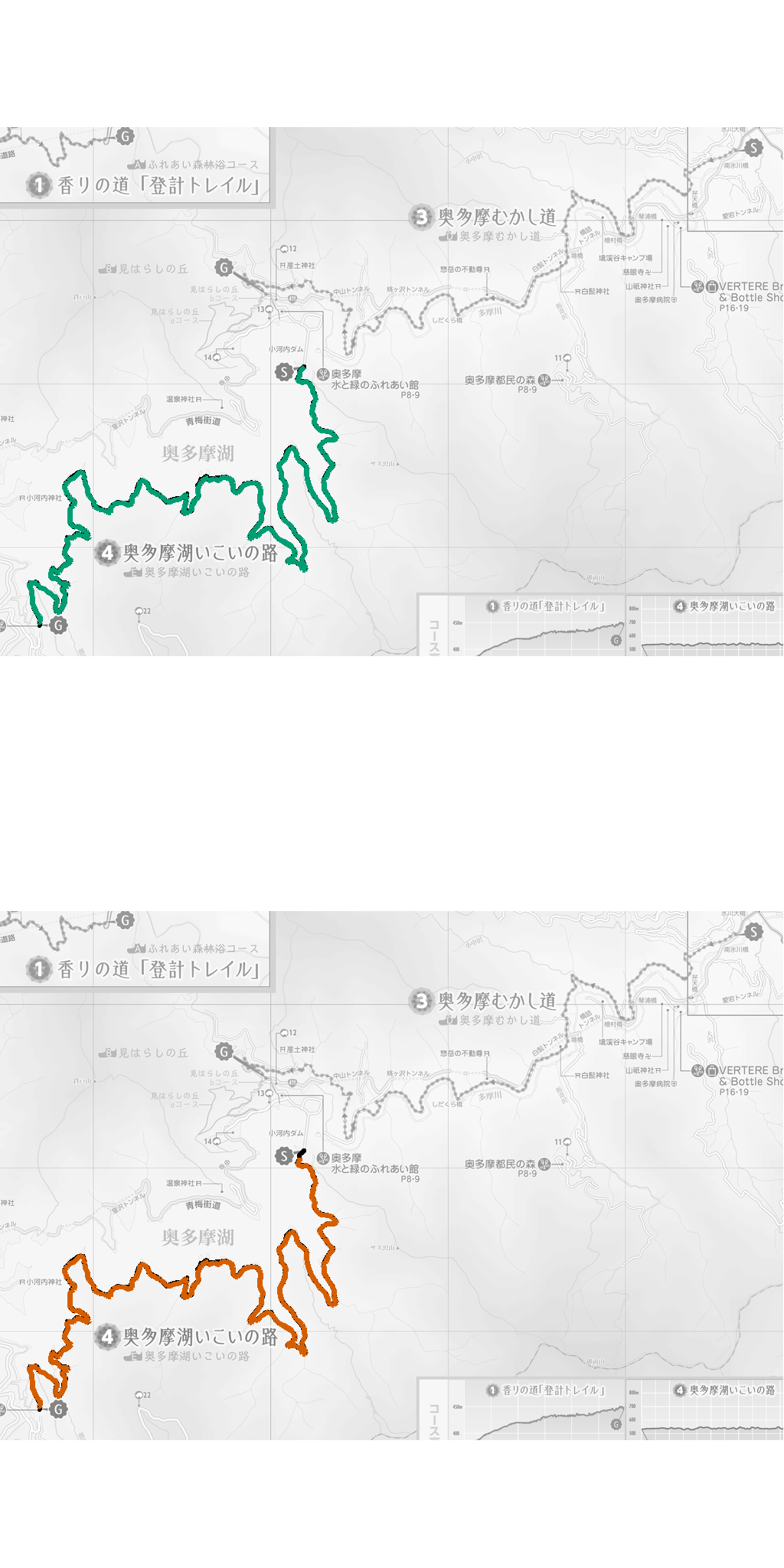}
        \caption{Ideal case: Chamfer distance of 8.7m (GT) and 8.94m (Pred)}
    \end{subfigure}
    \hfill
    \begin{subfigure}[t]{0.23\textwidth}
        \includegraphics[width=\linewidth]{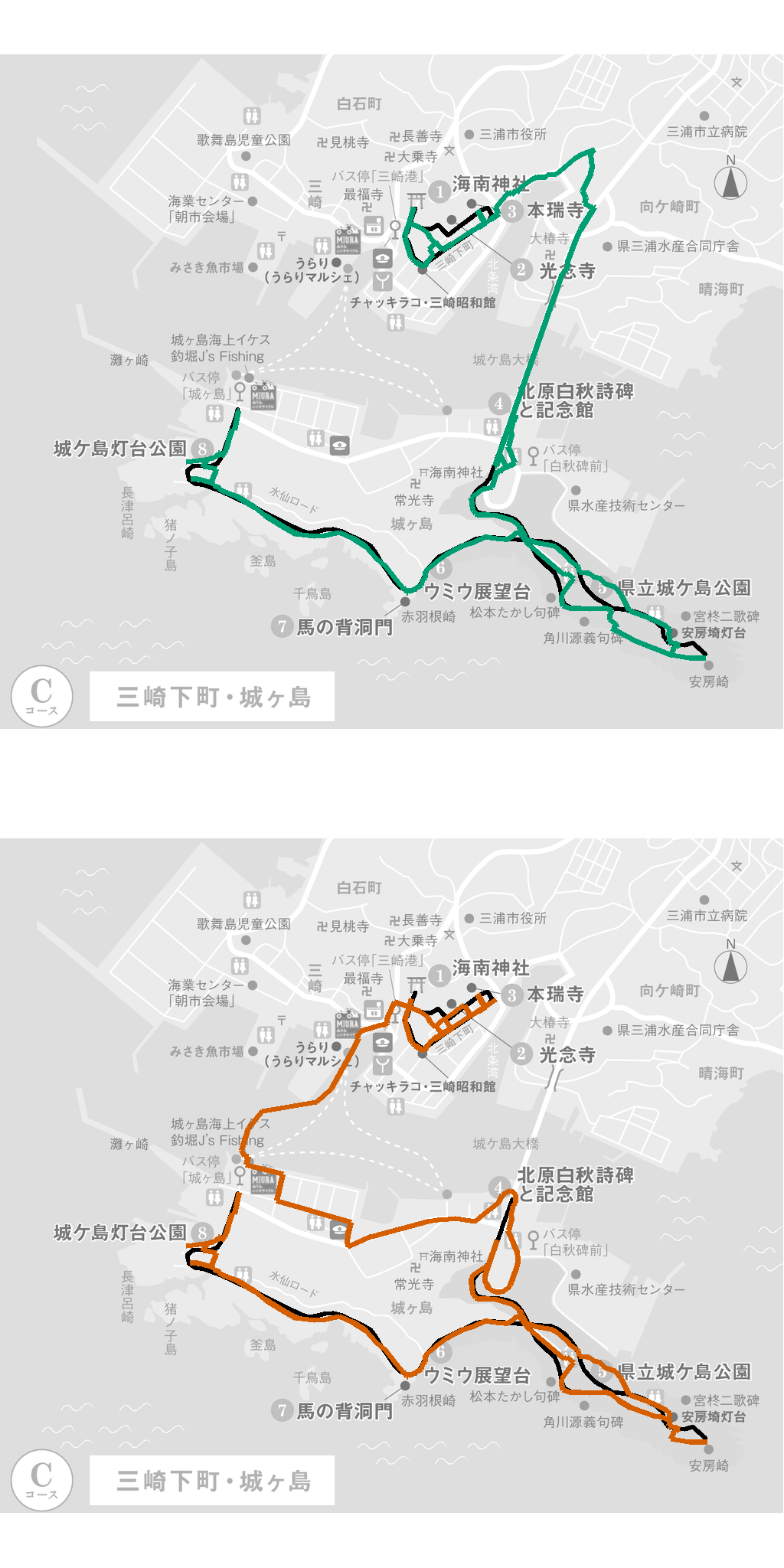}
        \caption{Average case 1: Chamfer distance of 26.77m (GT) and 19.15m (Pred)}
    \end{subfigure}
    \hfill
    \begin{subfigure}[t]{0.23\textwidth}
        \includegraphics[width=\linewidth]{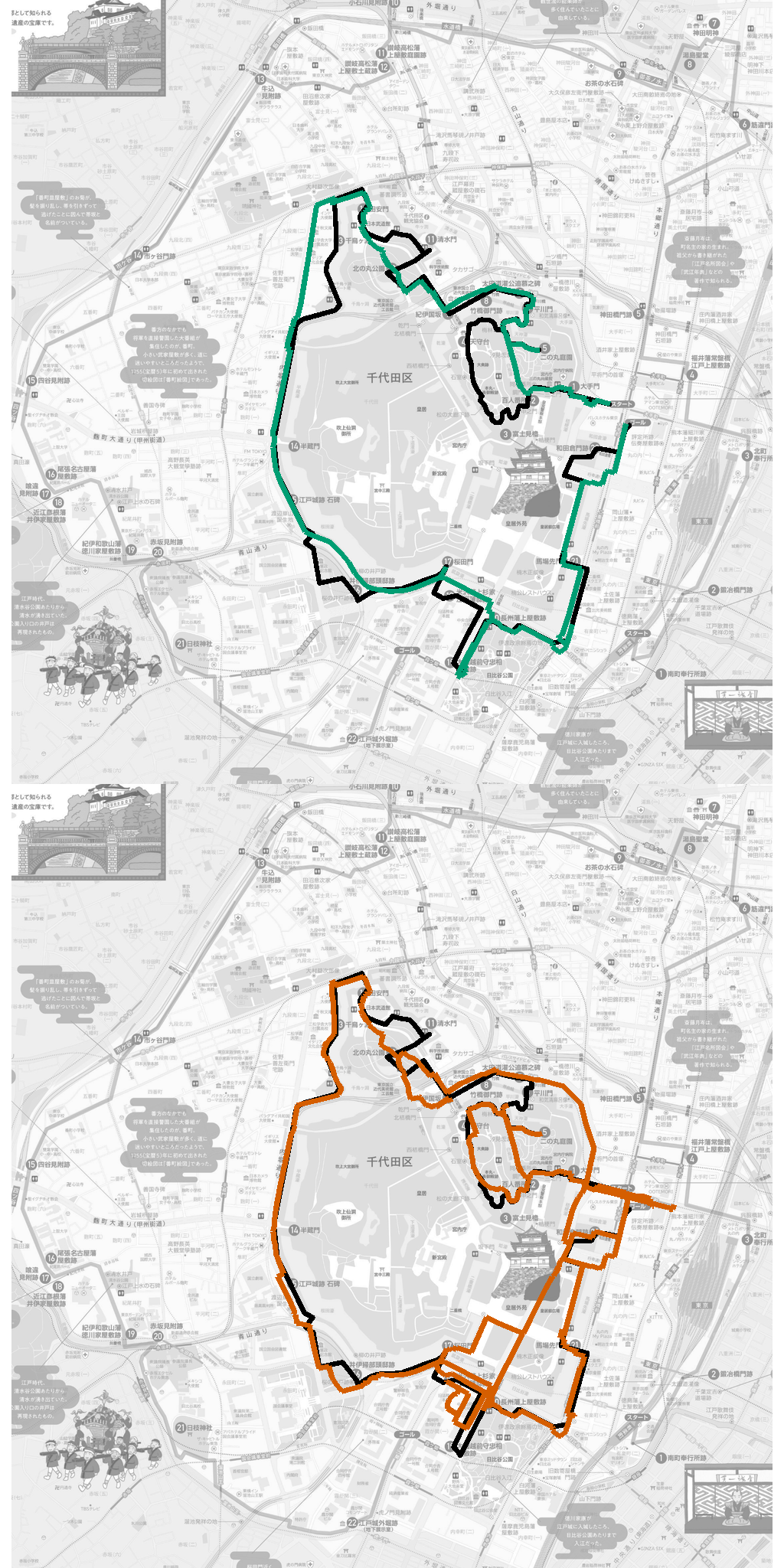}
        \caption{Average case 2: Chamfer distance of 34.21m (GT) and 29.68m (Pred)}
    \end{subfigure}
    \hfill
    \begin{subfigure}[t]{0.23\textwidth}
        \includegraphics[width=\linewidth]{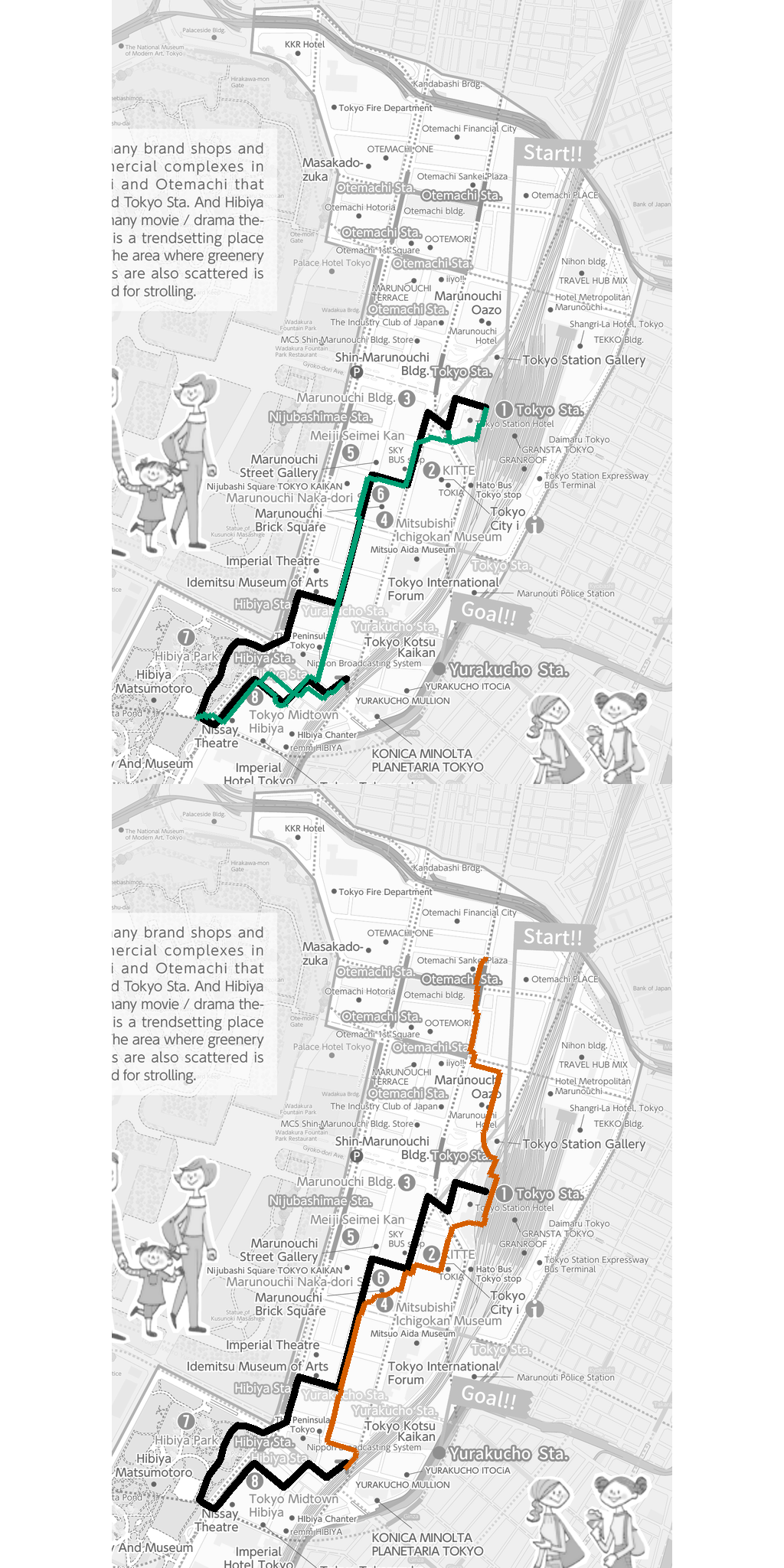}
        \caption{High error case: Chamfer distance of 39.35m (GT) and 151.77m (Pred)}
    \end{subfigure}
    \caption{Examples of generated routes. Each column shows the ground truth trail mask in black, overlaid on the corresponding grayscale input map. The top row shows the routes generated using ground truth masks as input (GT, green), and the bottom row shows routes generated using predicted masks (Pred, orange). The examples illustrate how noise in the predicted masks can lead to generated routes that cover areas not intended to be part of a trail.}
    \label{fig:route_outputs}
\end{figure*}

\subsection{Full Pipeline}

We now evaluate the full end-to-end pipeline by using the predicted trail masks as inputs to the route generation algorithm. For evaluation, the generated GPX routes are projected back into image space and compared to the ground truth trail masks.

To assess the alignment between GPX routes and trail masks, we report three distance-based metrics:
\begin{itemize}
    \item \textbf{GPX-to-Trail distance}: the average distance from each GPX point to the nearest pixel in the trail mask. Higher values indicate that the generated route deviates from the intended trail.
    \item \textbf{Trail-to-GPX distance}: the average distance from each trail mask pixel to the closest point of the GPX route. Lower values indicate better trail coverage by the route.
    \item \textbf{Chamfer distance}: the mean of the two distances above, providing an overall similarity score.
\end{itemize}

\cref{fig:distances} compares these distances for runs using ground truth versus predicted trail masks. As expected, performance degrades when using predicted masks. The GPX-to-Trail distance increases substantially, suggesting the route sometimes diverges from the true trail. The Trail-to-GPX distance is also higher, though the difference is smaller.

Interestingly, the interquartile range of the Trail-to-GPX distance is lower for the predicted-mask configuration, indicating less variability in performance.

The increased Chamfer distances are largely driven by the high GPX-to-Trail distances. These in turn appear to be usually caused by the segmentation model including wrongly detected regions in the mask. These false positives are mistakenly interpreted as valid trail segments and incorporated into the routing graph, leading the algorithm to route through areas that are not part of the actual trail. This can be observed in some of the example outputs visualized in \cref{fig:route_outputs}.

\begin{table*}[h]
\centering
\caption{Ablation study of initial query strategy and iterative refinement. All configurations are evaluated using Chamfer distance (meters, mean ± std).}
\label{tab:ablation}
\begin{tabular}{lcccccl}
\toprule
ID & Input & \makecell{End-to-End\\Query (E2E)} & \makecell{Graph\\Processing (GP)} & \makecell{Iterative\\Refinement (IR)} & \makecell{Strategy\\Description} & Chamfer ↓ \\
\midrule
A & GT & \checkmark & × & \checkmark & E2E + IR & 38.92 ± 38.58 \\
B & GT & × & \checkmark & \checkmark & GP + IR & \textbf{18.22 ± 11.14} \\
C & GT & \checkmark & \checkmark & \checkmark & Hybrid + IR & 18.25 ± 11.40 \\
D & GT & \checkmark & \checkmark & × & Hybrid (no IR) & 122.83 ± 105.92 \\
\midrule
E & Pred & \checkmark & × & \checkmark & E2E + IR & 56.10 ± 55.83 \\
F & Pred & × & \checkmark & \checkmark & GP + IR & \textbf{44.40 ± 47.00} \\
G & Pred & \checkmark & \checkmark & \checkmark & Hybrid + IR & 45.15 ± 48.67 \\
H & Pred & \checkmark & \checkmark & × & Hybrid (no IR) & 137.12 ± 115.32 \\
\bottomrule
\end{tabular}
\end{table*}

\subsection{Ablation Study}

\cref{tab:ablation} summarizes the configurations evaluated in our ablation study. Configurations A–D use ground truth trail masks as input to the pipeline (Input: GT), while configurations E–H use predicted trail masks produced by our segmentation model (Input: Pred).

To assess the impact of the initial query strategy, we compare configurations A–C and E–G, all of which include iterative refinement. For both GT and Pred inputs, the lowest Chamfer distances are achieved when using the graph-based query as the initial strategy. While the hybrid approach occasionally yields better results on individual samples, its overall median Chamfer distance is slightly higher although statistically insignificant for both GT and Pred inputs (\( \alpha = 0.05 \)).

To isolate the effect of the iterative refinement procedure, we compare the best-performing configurations with and without refinement. With refinement enabled, the optimal strategy uses the graph-based query (configurations B and F for GT and Pred, respectively). Without refinement, the best results are obtained by the hybrid strategy, which selects the better result between the end-to-end and graph-based queries (configurations D and H). As shown in \cref{tab:ablation}, iterative refinement consistently reduces Chamfer distance, demonstrating its effectiveness in improving route alignment under both ground truth and predicted mask conditions.

% Discussion
\section{Discussion and Future Work}

A major challenge in trail extraction from scanned maps is the stylistic diversity across different map sources. There are no consistent conventions among creators of hiking and sightseeing maps: trail representations can vary in color, stroke style, and overlap with other map elements, such as icons or textboxes. This diversity complicates segmentation and can lead to false positives, as observed in several of our U-Net predictions. Potential improvements include adjusting the classification threshold or incorporating contextual cues, such as detecting and interpreting map legends, to help the model better distinguish trail features.

\begin{figure}[t]
    \centering
    \begin{subfigure}[t]{0.48\linewidth}
        \includegraphics[width=\linewidth]{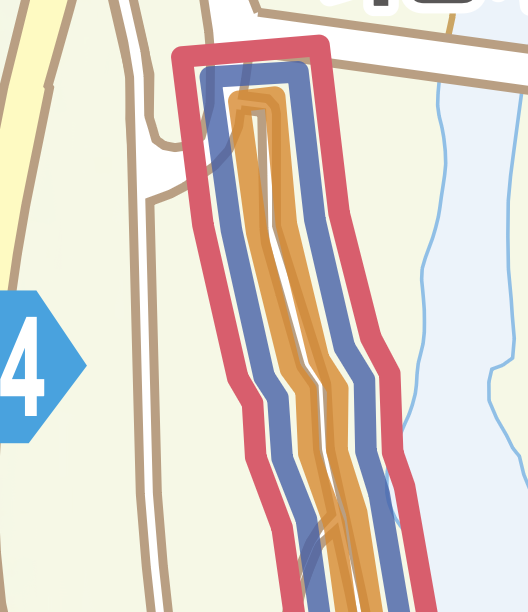}
        \caption{Multiple trails follow the same (narrow) street, leading to the trails being drawn well beyond the edges of the street.}
        \label{fig:misalignment_1}
    \end{subfigure}
    \hfill
    \begin{subfigure}[t]{0.48\linewidth}
        \includegraphics[width=\linewidth]{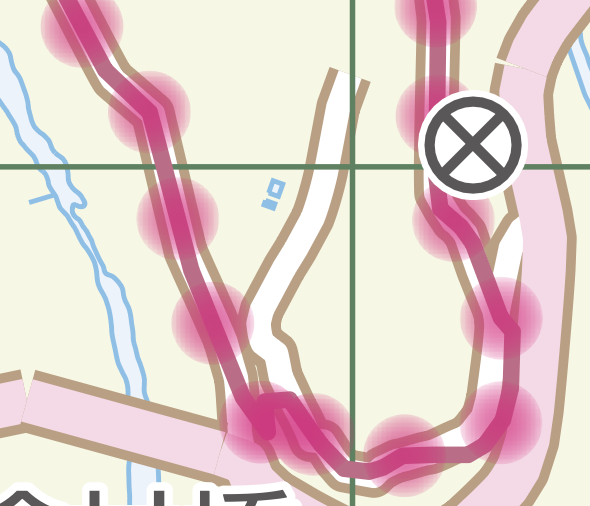}
        \caption{The trail is accidentally drawn to extend into a different street (right side).}
    \end{subfigure}
    \caption{Examples of misaligned trails in scanned maps. Trail keypoints may be manually placed, leading to inconsistencies with the basemap.}
    \label{fig:misalignment}
\end{figure}

Another source of complexity is that trails on many maps are drawn by manually placing keypoints over basemaps. This process can lead to visible misalignments between trails and the underlying topography (see \cref{fig:misalignment}). Such discrepancies support our decision to use a routing engine rather than directly sampling points from the trail mask, since even with perfect georeferencing, sampled points could fall off the actual path. Using a routing engine ensures that the generated route aligns with path and street data from OpenStreetMap. However, this introduces a structural mismatch: even when the generated route is semantically correct, it may not align pixel-perfectly with the trail mask when projected back into image space, particularly if the drawn trail was misaligned to begin with. A potential solution is to construct ground truth GPX routes and compare them directly to the generated routes, rather than relying solely on mask-based evaluation.
 
While most of our pipeline is fully automated, one step remains manual: providing GCPs for georeferencing. This currently limits the real world applicability of our pipeline, and integrating approaches for automatically finding GCPs is a logical next step in our research.

Finally, we note that the problem of route extraction from scanned maps remains largely unexplored. While we have shown that it is possible to extract usable routes from scanned maps with our pipeline, alternative approaches may further improve efficiency or accuracy. In particular, the graph-based processing stage is quite resource-intensive. Machine learning models could potentially accelerate the identification of key points for initial queries and might potentially even be combined with the segmentation step in the future.

{
    \small
    \bibliographystyle{ieeenat_fullname}
    \bibliography{main}
}

\end{document}